\documentclass[11pt,a4paper]{article}
\usepackage[hyperref]{emnlp-ijcnlp-2019}
\usepackage{times}
\usepackage{latexsym}
\usepackage{graphicx}
\usepackage{lscape}
\usepackage{hhline}
\usepackage{multirow}
\usepackage{amsmath}
\usepackage[ruled,longend]{algorithm2e}
\definecolor{orange}{RGB}{255,127,0}
\definecolor{darkgreen}{RGB}{0,51,0}
\usepackage{tabularx}
\usepackage{subcaption}

\usepackage{url}

\aclfinalcopy 

\title{Cracking the Contextual Commonsense Code: Understanding Commonsense Reasoning Aptitude of Deep Contextual Representations}

 \author{
    Jeff Da\quad \quad \quad \quad
	Jungo Kasai\\
   Paul G.~Allen School of Computer Science \& Engineering,\\University of Washington, Seattle, WA, USA \\
    {\tt \{jzda,jkasai\}@cs.washington.edu}
}
\date{}

\begin{document}
\maketitle
\begin{abstract}
Pretrained deep contextual representations have advanced the state-of-the-art on various commonsense NLP tasks, but we lack a concrete understanding of the capability of these models. Thus, we investigate and challenge several aspects of BERT's commonsense representation abilities. First, we probe BERT's ability to classify various object attributes, demonstrating that BERT shows a strong ability in encoding various commonsense features in its embedding space, but is still deficient in many areas. Next, we show that, by augmenting BERT's pretraining data with additional data related to the deficient attributes, we are able to improve performance on a downstream commonsense reasoning task while using a minimal amount of data. Finally, we develop a method of fine-tuning knowledge graphs embeddings alongside BERT and show the continued importance of explicit knowledge graphs.
\end{abstract}

\section{Introduction}

Should I put the toaster in the oven? Or does the cake go in the oven? Questions like these are trivial for humans to answer, but machines have a much more difficult time determining right from wrong. Researchers have chased mimicking human intelligence through linguistic commonsense as early as \citet{mcc}:

\begin{quote} 
... [machines that] have much in common with what makes us human are described as having common sense. \cite{mcc}.
\end{quote}
Such commonsense knowledge presents a severe challenge to modern NLP systems that are trained on a large amount of text data.
Commonsense knowledge is often implicitly assumed, and a statistical model fails to learn it by this reporting bias \cite{Gordon2013ReportingBA}.
This critical difference of machine learning systems from human intelligence hurts performance when given examples outside the training data distribution \cite{Gordon2013ReportingBA,Schubert2015WhatKO,Davis2015CommonsenseRA,Sakaguchi2019WINOGRANDEAA}. 

On the other hand, NLP systems have recently improved dramatically with contextualized word representations in a wide range of tasks \cite{Peters2018, openaigpt, devlin2018}.
These representations have the benefit of encoding context-specific meanings of words that are learned from large corpora.
In this work, we extensively assess the degree to which these representations encode grounded commonsense knowledge, and investigate whether contextual representations can ameliorate NLP systems in commonsense reasoning capability.

We present a method of analyzing commonsense knowledge in word representations through attribute classification on the semantic norm dataset \cite{Devereux2014}, and compare a contextual model to a traditional word type representation.
Our analysis shows that while contextual representations significantly outperform word type embeddings, they still fail to encode some types of the commonsense attributes, such as visual and perceptual properties. In addition, we underscore the translation of these deficiencies to downstream commonsense reasoning tasks.

We then propose two methods to address these deficiencies: one implicit and one explicit. Implicitly, we train on additional data chosen via attribute selection. Explicitly, we add knowledge embeddings during the fine-tuning process of contextual representations. This work shows that knowledge graph embeddings improve the ability of contextual embeddings to fit commonsense attributes, as well as the accuracy on downstream reasoning tasks.

%
%
%
%
%
%

\section{Attribute Classification}

\begin{figure*}[t]
    \centering
    \begin{subfigure}[t]{0.48\textwidth}
        \centering
          \includegraphics[width=0.98\textwidth]{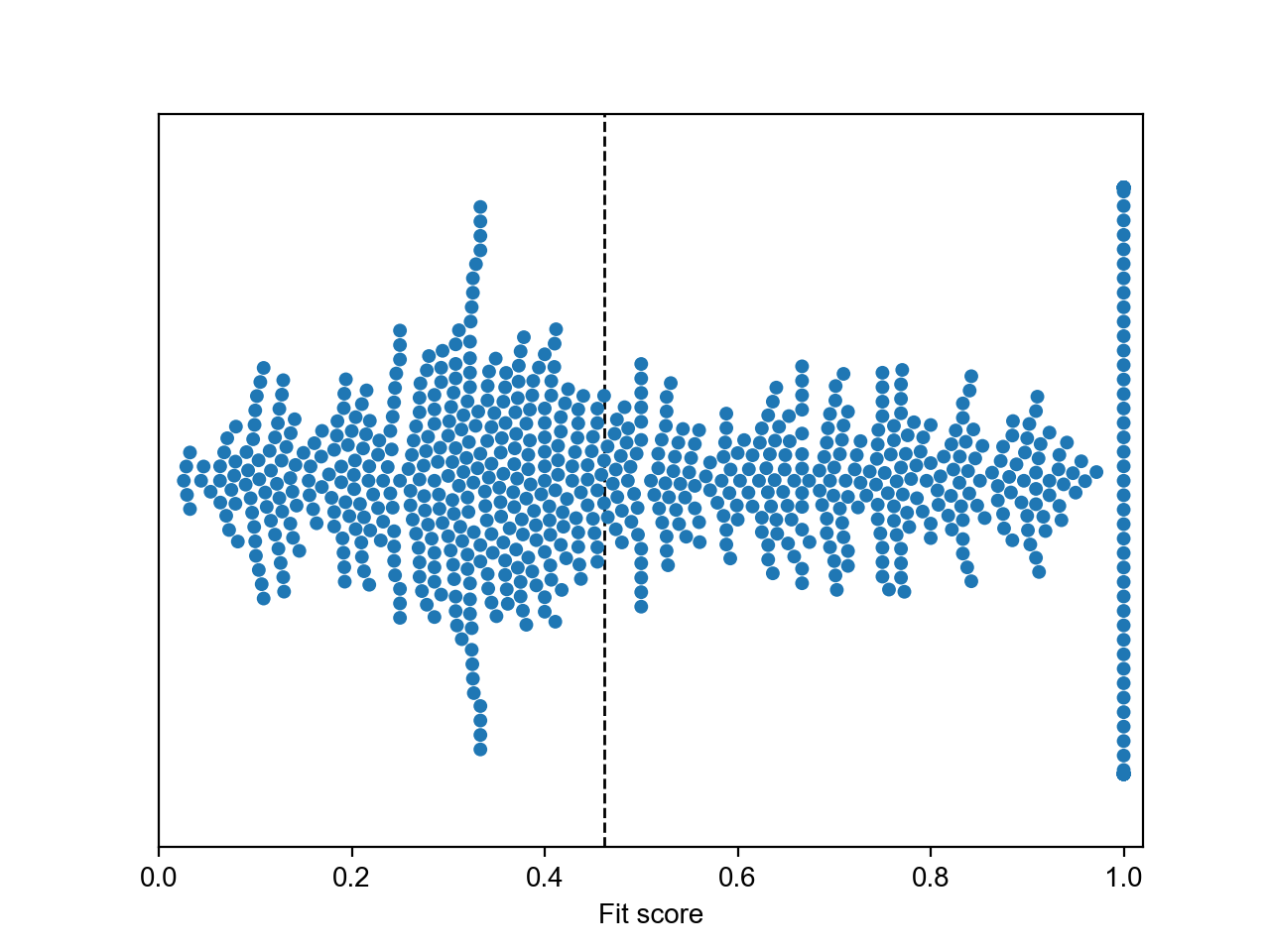}
        \caption{GloVe}
    \end{subfigure}%
    \begin{subfigure}[t]{0.48\textwidth}
        \centering
          \includegraphics[width=0.98\textwidth]{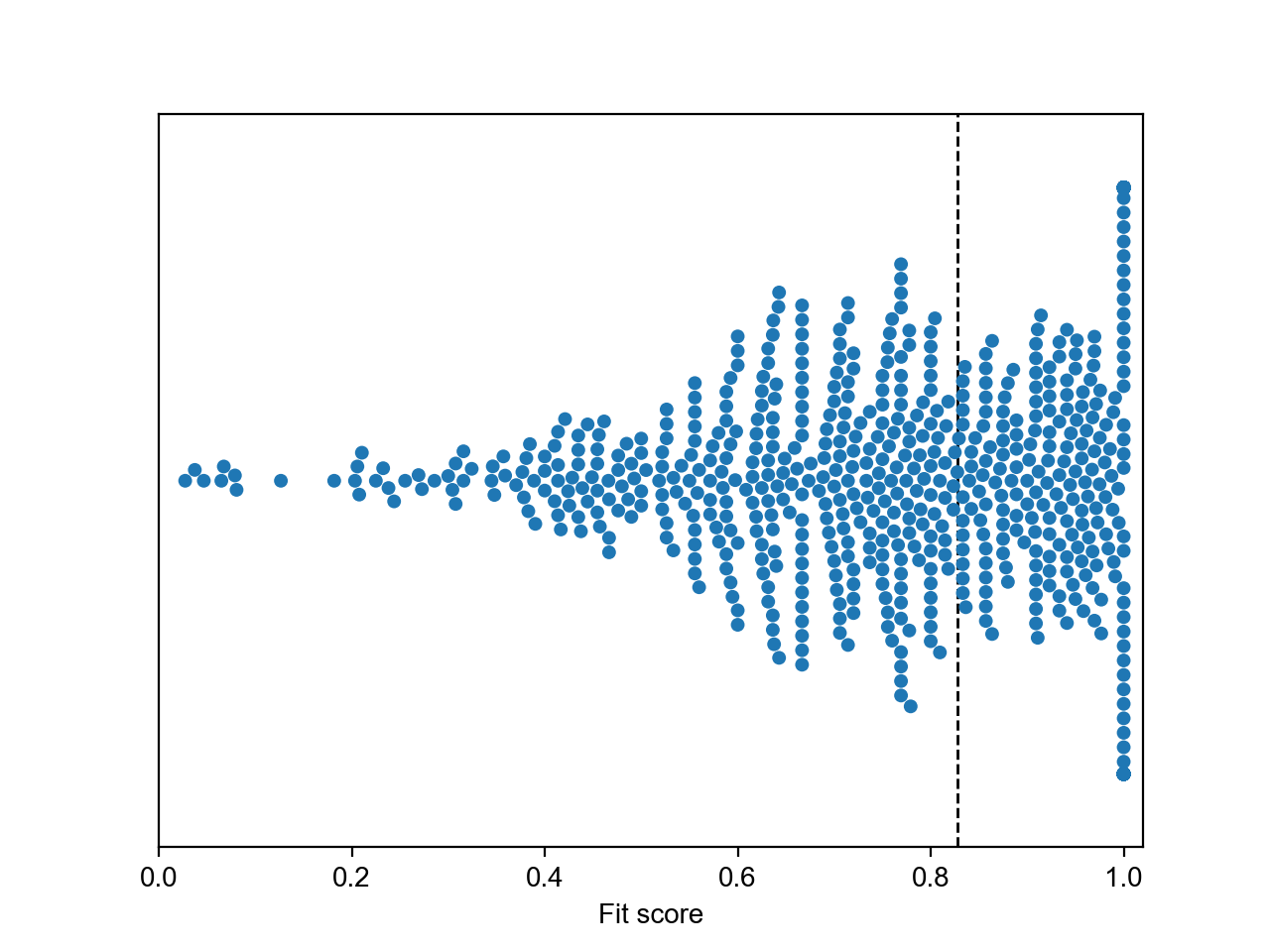}
        \caption{BERT}
    \end{subfigure}
    \caption{Swarm plots showing attribute fit scores for GloVe (left) and BERT (right). Each dot represents a single attribute, displayed along the x-axis according to the classifier's ability to fit that feature with the given embeddings. The y-axis is not significant, and instead, dots are displaced along the y-axis instead of overlapping to show quantity. The median fit score per embedding type is displayed with a dotted line.}
    \label{attribute_fit}
\end{figure*}

First, we preform an investigation to see if the output from BERT is able to encode the necessary features to determine if an object has a related attribute. We propose a method to evaluate BERT's representations and compare to previous non-contextual GloVe \cite{pennington-etal-2014-glove} baselines, using simple logistic classifiers.

\begin{figure*}[t]
    \centering
    \begin{subfigure}[t]{0.48\textwidth}
        \centering
          \includegraphics[width=0.98\textwidth]{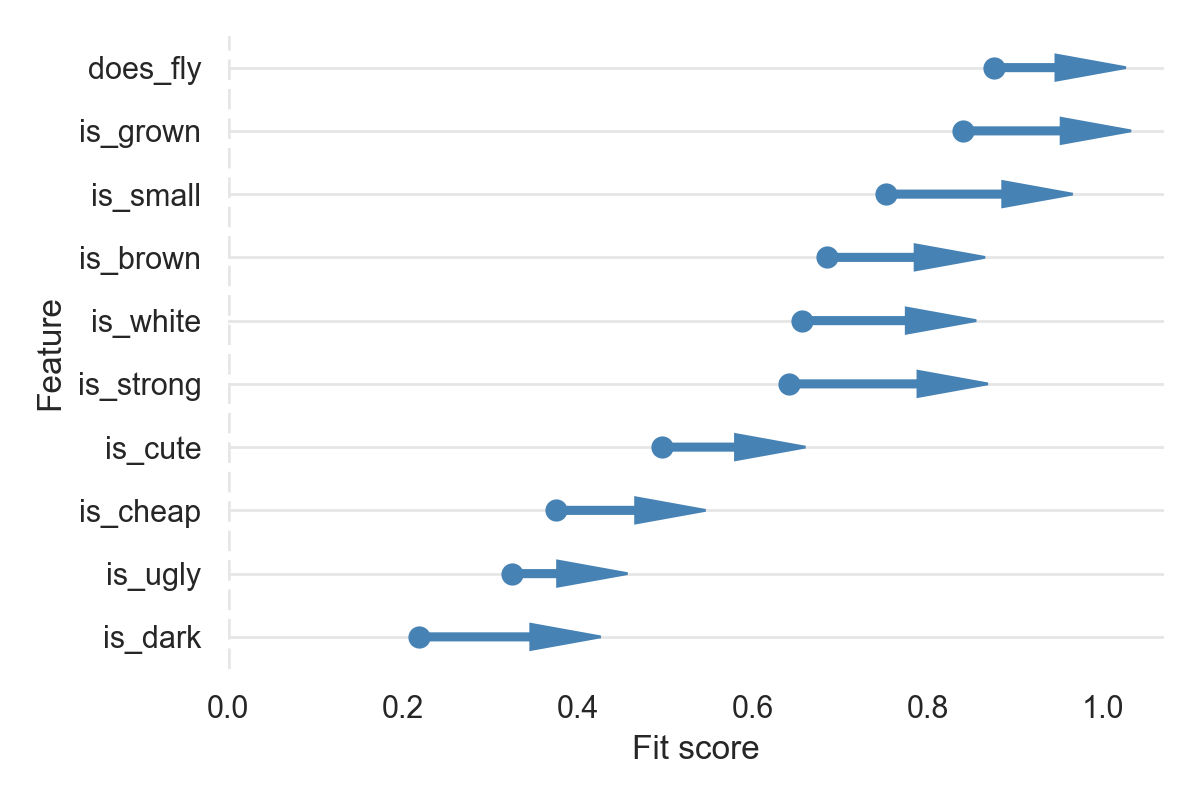}
        \caption{Small increase in fit score ($<$ 0.15)}
    \end{subfigure}%
    \begin{subfigure}[t]{0.48\textwidth}
        \centering
          \includegraphics[width=0.98\textwidth]{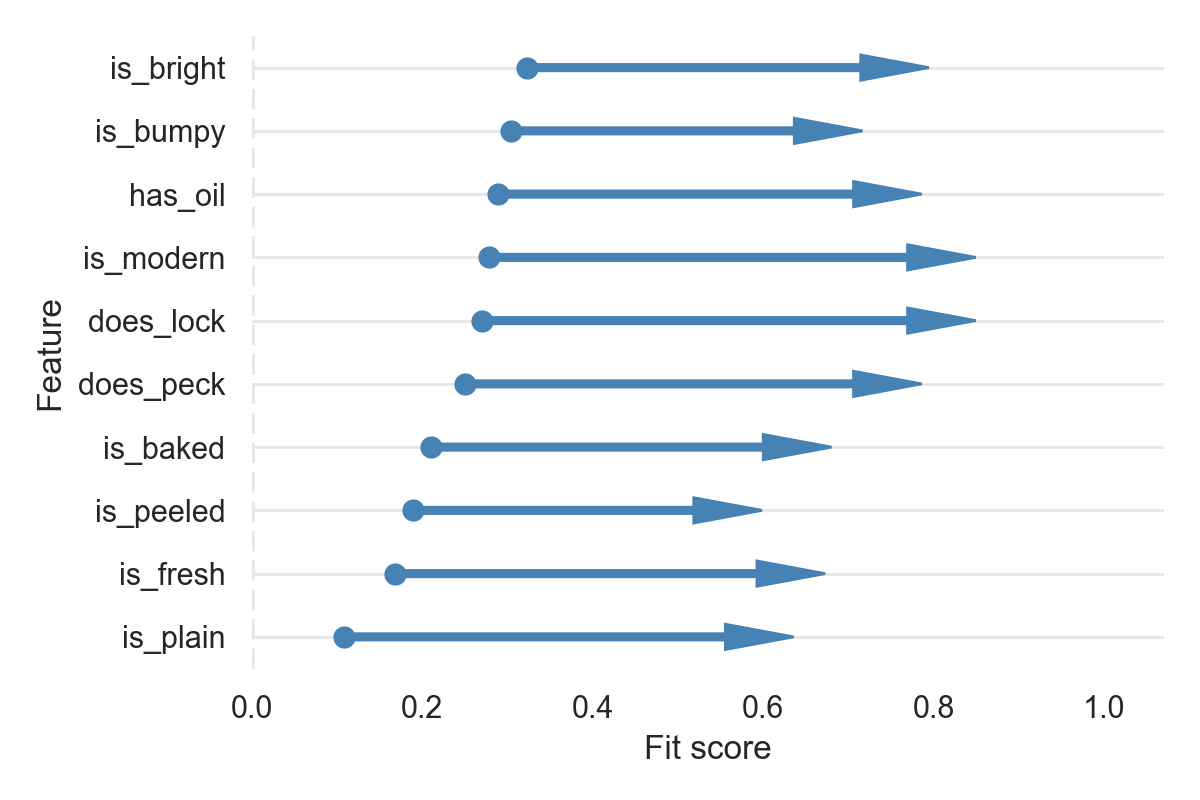}
        \caption{Large increase in fit score ($>$ 0.3)}
    \end{subfigure}
    \caption{Differences between fit scores when using GloVe (start of arrow) or BERT (end of arrows) embeddings.}
    \label{arrowgraph}
\end{figure*}
\subsection{Commonsense Object Attribution}

To get labels for attribute features of commonsense features of objects, we utilize CSLB, a semantic norm dataset collected by the Cambridge Centre for Speech, Language, and the Brain \citep{Devereux2014}. Semantic norm datasets are created through reports from human participants asked to label the semantic features of a given object. Thus, a proportion of these features are obvious to humans, but may be difficult to find written in text corpora. This is notably different from the collection methods of prominent commonsense databases, such as ConceptNet \cite{conceptnet}.

CSLB gives 638 different attributes describing a variety of objects provided by 123 participants. To make results consistent between baselines (GloVe) and BERT, we first preprocess the attributes present in CSLB. We removed attributes with two-word names, ambiguous meanings (i.e. homographs), or missing GloVe representations. This gives a 597 attribute vocabulary. Examples of objects described are \textit{zebra}, \textit{wheel}, and \textit{wine}.
Example of attributes are \textit{is upright}, \textit{is a toy}, and \textit{is an ingredient}.


\subsection{Contextualization}

Since BERT is commonly utilized at the sequence embedding level \cite{devlin2018}, we develop a contextualization module to allow representations of (\textit{object, attribute}) pairs, allowing us to acquire one sequence embedding from BERT for each pair. From a high level, we want to develop a method to transform (\textit{object, attribute}) into simple grammatical sentences.

For each \textit{(object, attribute)} pair, we raise the pair to a sentence structure such that the attribute is describing the object. We would enforce the following representation, in line with the procedure of \citet{devlin2018}:

$[CLS]$ $c_{\text{prefix}}$ noun $c_{\text{affix}}$ adj. $c_{\text{postfix}}$ $[SEP]$

The goal is to create a simple formula that allows the model to isolate the differences between the object-attribute (noun-adjective) pairs, rather than variation in language.  $c_{\text{prefix}}$ represents previous context, i.e.\ context that appears before the word. $c_{\text{affix}}$ is context that appears between the noun and the adjective. $c_{\text{postfix}}$ is context that closes out the sentence.

We illustrate this algorithm for use with CSLB, but this methodology can be used for any dataset, such as other semantic norm datasets. We use this process for each \textit{(object, attribute)} pair in CSLB. First, we check if any words in the attribute need to be changed. For example, in CSLB, instead of \textit{does deflate}, we use \textit{deflates} as the attribute text, since it simplifies the language. Then, for $c_{\text{prefix}}$, we use either \textit{A} or \textit{An}, and for $c_{\text{postfix}}$, and use a period. For $c_{affix}$, we use either \textit{is} or nothing, depending on the attribute. Some example sentences would be: \textit{(shirt, made of cotton)} would become "A shirt is made of cotton." and \textit{(balloon, does deflate)} becomes "A balloon deflates." See the appendix for full pseudocode.

We find that this method is a better alternative to simply creating a sequence with the concatenation of the object and the attribute. Some attribute-object pairs translate better to English than others. For example, "wheel does deflate" might be a relatively uncommon and awkward English phrase when compared to more natural phrases such as "shirt made of cotton".

\subsection{Determining Attribute Fit}

We explore if word embeddings contain the necessary information within their embedding space to classify various semantic attributes. Our procedure involves use of a simple logistic classifier to classify if an \textit{attribute} applies to a candidate \textit{object}. We create a list of \textit{(object, attribute)} pairs as training examples for the logistic classifier (thus, there are $n_{objects} \times n_{attributes}$ training examples in total). We then train logistic classifiers for each attribute, and use leave-one-out accuracy as accuracy -- averaging the leave-out-one result across all $n_{objects}$ classifiers, since we leave out a different object each time. For example, to examine the attribute \textit{made of cotton}, we train on all objects except one, using the label $1$ if the object is made of cotton, and $0$ otherwise. Then, we test to see if the left-out object is classified correctly. We repeat $n_{objects}$ times, removing a different object each time. To judge fit, we use F1 score, as F1 score is not affected by dataset imbalance. We consider other classifiers, such as SVD classifiers, but we find that there is no significant empirical difference between the classifiers. For baseline tests, we use the pretrained 300 dimensional GloVe embeddings,\footnote{\url{https://nlp.stanford.edu/projects/glove/}} as they have shown to perform better than word2vec embeddings \citep{lucy2017}. See appendix for specific logistic regression parameters, such as the number of update steps used.

\subsection{Attribute Scores}
\label{attscores}

\begin{table*}
\small
\resizebox{\textwidth}{!}{%
\begin{tabular}{lccccc||c}
\hline
Metric & Visual   & Encyclopedic & Functional & Perceptual & Taxonomic & Overall   \\ \hline
Median$_{GloVe}$     & 46.2    & 38.9         & 44.4      & 49.0      & 89.1     & 46.1     \\[1pt]
Median$_{BERT}$     & 83.3   & 76.2         & 78.3      & 80.0     & 100      & 82.7    \\[1pt] \hline
$\Delta$ & \textbf{+37.1} & \textbf{+37.3} & \textbf{+33.9} & \textbf{+31.0} & \textbf{+10.9} & \textbf{+36.6} \\ \hline
\end{tabular}%
}
\caption{Comparison of median logistic classifier fit scores (out of 100 percent fit) across categories defined in CSLB.}
\label{results_categories}
\end{table*}

\begin{table*}[]
\begin{tabularx}{\textwidth}{lXX} \hline
Category & Lower scoring attributes (fit score \textless$\ $1.0) & Attributes perfectly fit (fit score = 1.0) \\ \hline
Visual & is triangular, is long and thin, is upright, has two feet, does swing, is rigid & does come in pairs, has a back, has a barrel, has a bushy tail, has a clasp \\
Encyclopedic & is hardy, has types, is found in bible, is American, does play, is necessary essential & does grow on plants, does grow on trees, does live in rivers, does live in trees, does photosynthesize, has a crew \\
Functional & does work, does spin, does support, does drink, does breathe, does hang & does DIY, does carry transport goods, does chop, does drive \\
Perceptual & is chewy, does rattle, is wet, does squeak, is rough, has a strong smell & does bend, has a sting, has pollen, has soft flesh, is citrus, is fermented \\
Taxonomic & is a home, is a dried fruit, is a garden tool, is a vessel, is a toy, is an ingredient & is a bird of prey, is a boat, is a body part, is a cat, is a citrus fruit, is a crustacean \\ \hline
\end{tabularx}
\caption{Fine-grained comparison across categories between attributes that lack some level of fit (left) and perfectly fit attributes (right) with classification using BERT representations.}
\label{examples}
\end{table*}

We show our findings for feature fit for each attribute. Figure \ref{attribute_fit} highlights that BERT is much stronger on this benchmark -- the median fit score is nearly double that of the previously reported GloVe baselines. This suggests that BERT encodes commonsense traits much better than previous baselines, which is suggestive of its strong scores on several commonsense reasoning tasks.
Notably, we can see that much fewer features have a fit score less than 0.5. 
We observe that many more traits have a perfect fit score of 1.0. However, our results also show that BERT is still unable to fully fit many attributes. This underscores that BERT still lacks much attribution ability, perhaps in areas outside of its training scheme or pretraining data.
Seen in Figure \ref{arrowgraph} is the change in fit scores between GloVe and BERT. We can see that some traits exhibit much larger increases -- in particular, physical traits such as \textit{made of wood}, \textit{does lock,} and \textit{has a top}. Traits that are more abstract tend to have a lesser increase. For example, \textit{is creepy} and \textit{is strong} still are not able to be fit by the contextualized BERT module.

Table \ref{results_categories} shows a comparison of fit scores across different types of attribute categories. These categories are defined per attribute in CSLB \cite{Devereux2014}. Visual attributes define features that can be perceived visually, such as \textit{is curved}. Perceptual defines attributes that can be perceived in other non-visual ways, such as \textit{does smell nice}. Functional describes the ability of an object, such as \textit{is for weddings}. Taxonomic defines a biological or symbolic classification of an object like \textit{is seafood}. Finally, encyclopedic are traits that may be the most difficult to classify, as they are attributes that most pertain to abstract commonsense, such as \textit{is collectible}.

BERT has stronger scores in all categories, and just short of double the overall accuracy. Importantly, however, it struggles to classify many categories of objects. In taxonomic categories, it is able to perfectly fit more than half the objects. We suspect that this is intuitive, as BERT is trained on text corpora that allow for learning relationships between classes of objects and the object itself.
GloVe notably also preforms strong in this category, for the same reasons. BERT scores the lowest on encyclopedic traits, which most closely resemble traits that would appear in commonsense tasks.
This suggests that BERT maybe be relatively deficient in regards to reasoning about commonsense attributes.

We also examine specific attributes where BERT is fully fit (with a perfect fit score), and compare those attributes to features where BERT is unable to fit. Table \ref{examples} shows examples of both levels of fit. BERT is able to fit many features that would be easily represented in text, such as $does\ bend$, $does\ grow\ on\ plants$, and $does\ drive$. It is unable to fit traits that may be less common in text and more susceptible to the reporting bias, such as $is\ American$, $is\ chewy$, and $has\ a\ strong\ smell$. Surprisingly, it is also unable to fit several features that would be likely common in text such as $is\ a\ toy$, suggesting that BERT's training procedure is lacking coverage of many everyday events perhaps due to the reporting bias.

\subsection{Do Knowledge Graphs Help?}
\label{explicit}

We extend our investigation with two inquiries. First, given the large gain in accuracy over GloVe, we wonder if BERT embeddings now encode the same information that external commonsense knowledge graphs (such as ConceptNet \cite{conceptnet}) provide. Second, we question if it is possible to increase the overall accuracy above the accuracy presented by using BERT embeddings (otherwise, it could mean that the deficit is simply because the logistic classifier does not have needed capacity \cite{Liu2019LinguisticKA}).

We use ConceptNet \cite{conceptnet} for our experiments. We label each relationship type with an index. ($antonym$ as $0$, $related\_to$ as $1$, etc.) During classification, we query the knowledge base with the object and the attribute and check if there are any relationships between the two. We embed the indexes of matched relationships to randomly initialized  embeddings and concatenate them with the original BERT embeddings. If more than one relationship is found, we randomly choose a relationship to use.

\begin{table}[]
\centering
\begin{tabular}{ll} \hline
System & Median \\ \hline
GloVe & 46.1 \\
BERT$_{LARGE}$ & 82.7 \\
ConceptNet & 23.2 \\
BERT$_{LARGE}$ + ConceptNet & \textbf{90.7} \\ \hline
\end{tabular}
\caption{Results for attribute classification with ConceptNet as a knowledge graph source.}
\label{conceptnetatt}
\end{table}

Table \ref{conceptnetatt} shows our results. By itself, the explicit commonsense embeddings do not have enough coverage to learn classifications of each attribute, since the knowledge graph does not contain information about every $(object$, $attribute)$ pair. However, by combining the knowledge graph embeddings with the BERT embeddings, we illustrate that knowledge graphs cover information that BERT is unable to generate the proper features for. In addition, the results suggest that BERT is deficient over various attributes, and the traditional knowledge graphs are able to cover this feature space. These results support the hypothesis that BERT simply lacks the features rather than the problem of the logistic classifier. 

\section{Improving BERT's Representations}

We have gained an understanding of the types of commonsense attributes BERT is able to classify and encode in its embeddings, and also have an understanding of the types of attributes that BERT's features are deficient in covering. In Section \ref{explicit}, we have shown that commonsense knowledge graphs may also help encode information that extends beyond BERT's embedding features. However, we have yet to know whether this BERT's deficiency will translate to any of BERT's downstream reasoning ability, which is ultimately more important.

\begin{figure*}
    \centering
  \includegraphics[width=\textwidth]{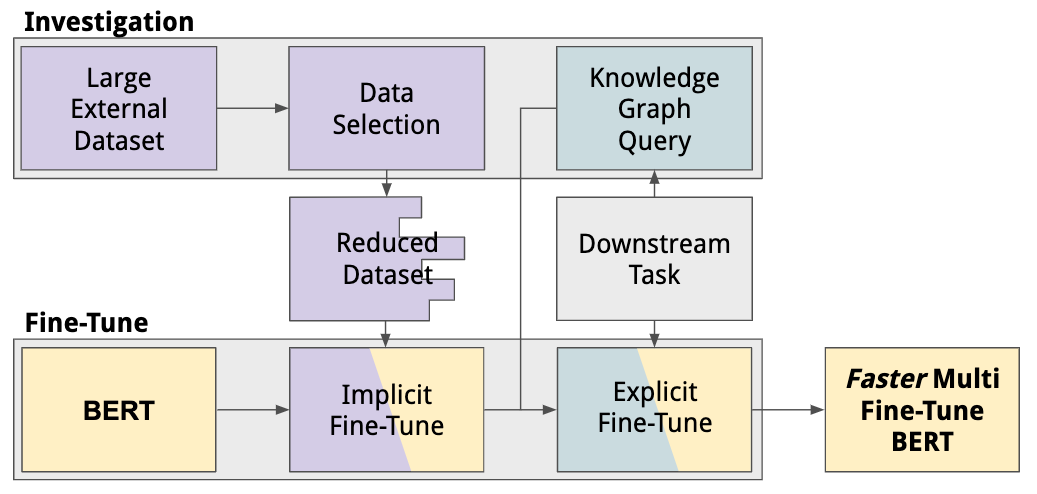}
\caption{Outline of our baseline method of improving BERT for commonsense reasoning. Our method fine-tunes BERT through multiple facets while optimizing for accuracy and reduced train steps. We use RACE \cite{Lai2017RACELR} as an external dataset, and MCScript 2.0 \cite{mcscript2} as our downstream task.}
\label{outline}
\end{figure*}

We empirically address the gap between attribute classification and downstream ability in BERT.
First, we demonstrate that there is a correlation between low-scoring attributes and low accuracy on reasoning questions that involve those attributes.
Then, we leverage our investigation to build two baseline methods of improving BERT's commonsense reasoning abilities (Figure \ref{outline}). Since BERT is trained on implicit data, we explore a method of using RACE \cite{Lai2017RACELR} alongside a list of attributes that BERT is deficient in (such as the one in Section \ref{attscores}).
We also extend our investigation in Section \ref{explicit} on commonsense knowledge graphs by proposing a method to integrate BERT with external knowledge graphs. See appendix for hyperparameters.

\subsection{Background: MCScript 2.0}

\begin{table}[]
\begin{tabularx}{0.46\textwidth}{X} \hline
Passage: For my anniversary with my husband, I decided to cook him a very fancy and nice breakfast. One thing I had always wanted to do but never got to try was making fresh squeezed orange juice. I got about ten oranges because I wasn't sure how much I was going to need to make enough juice for both me and my husband. I got home and pulled my juicer out from underneath my sink. I began using the juicer to squeeze the juice out of my orange juice. I brought my husband his breakfast with the orange juice, and he said that the juice was his favorite part! \\ \hline
How were the oranges sliced? \\
\textbf{a) in half}  \\
b) in eighths \\ \hline
When did they plug the juicer in? \\
a) after squeezing oranges \\
\textbf{b) after removing it from the box} \\ \hline
\end{tabularx}
\caption{Example of a prompt from MCScript 2.0 \cite{Ostermann2018MCScriptAN}, an everyday commonsense reasoning dataset. Questions often require script knowledge that extends beyond referencing the text.}
\label{examplemc}
\end{table}

We leverage MCScript 2.0 \cite{mcscript2} for several investigations in this paper. MCScript 2.0 is a downstream commonsense reasoning dataset. Each datum involves one passage, question, and two answers, and the goal is to pick the correct answer out of the two choices. Many questions involve everyday scenarios and objects, which helps us link our semantic norm results to more downstream reasoning capability. Table \ref{examplemc} shows an example.

\subsection{Do Low Classification Scores Result in Low Performance?}

\begin{figure}
    \centering
    \includegraphics[width=0.48\textwidth]{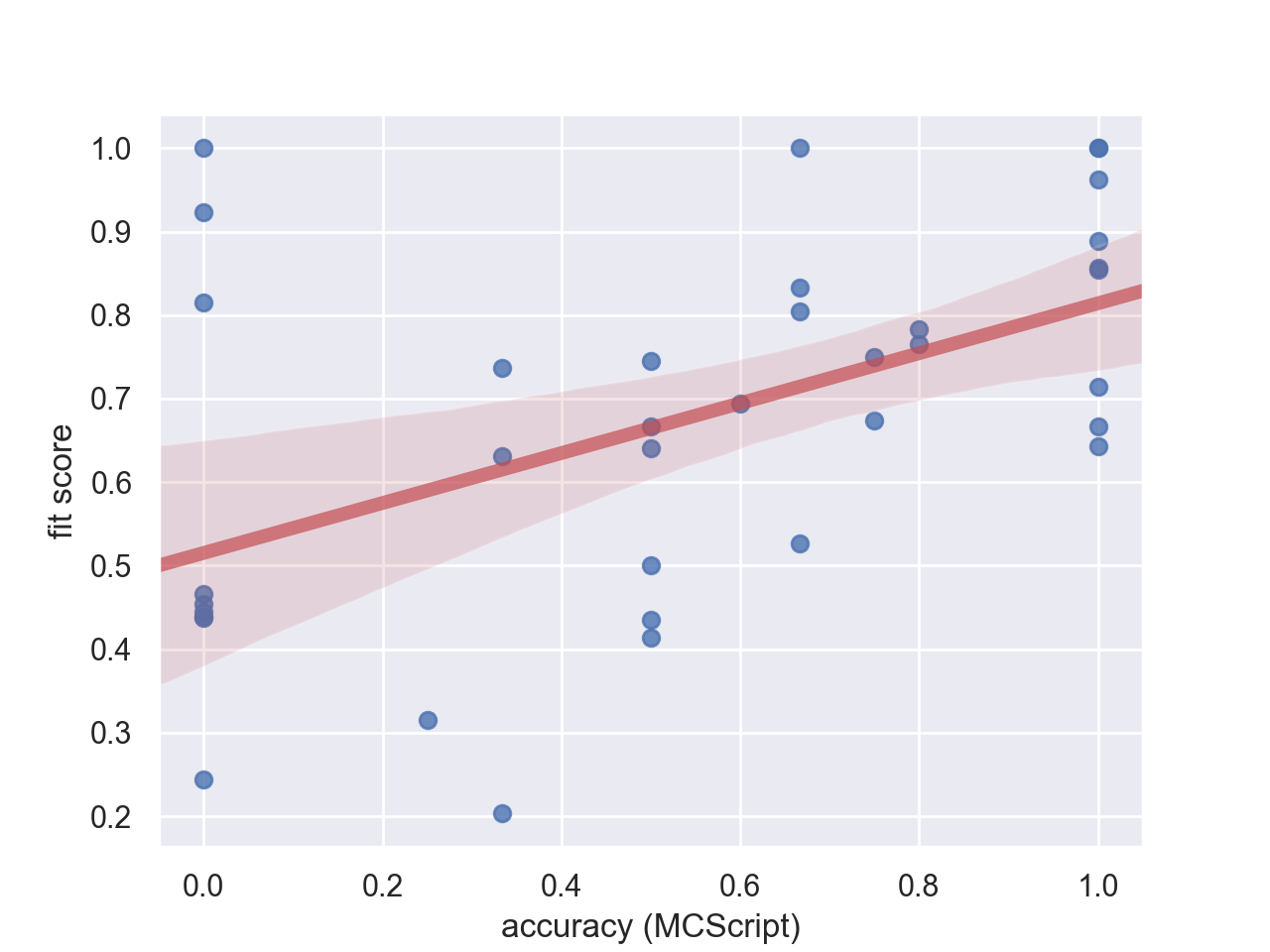}
    \caption{Linear regression fit of accuracy on MCScript 2.0, per attribute, versus fit score, with the inner 90 percent bootstrap confidence intervals highlighted (n = 1000). Each dot represents the accuracy of questions related to one attribute.}
    \label{mcscript_fit}
\end{figure}

We examine if low-scoring attributes result in low downstream performance, and high-scoring attributes also result in high downstream performance. For each question in MCScript, we relate that question to 1 or more of the attributes in the previous experiment. For example, a question might be talking about whether to use a camera flash, and would be thus related to the traits \textit{does have flash}, \textit{is dark}, and \textit{is light}.
Here we aim to empirically assess deficiencies in BERT's ability and their downstream implications. 
For instance, if it is unable to fit \textit{does have flash}, will it have a gap in knowledge in areas regarding camera flash? If a given feature does not have a related question, we do not include it in our experiments. In total, $n_{\text{questions}}$ = 193, and $n_{\text{attributes}}$ = 92.

For the MCScript model, we simply classify based on the $[CLS]$ token, as suggested in \citet{devlin2018}. We softmax over the logits between the two answers when producing our final answers, and split the passage-question pair and answer by a $[SEP]$ token. The attribute-related questions here are from the development set only.

Seen in Figure \ref{mcscript_fit} are the results.
We do not see a clear pattern, but we can still make several observations.
First, we notice that there are simply a lot of items with a high fit score. Next, there are a lot of attributes that BERT simply gets correct. However, notably, BERT is less consistent with getting items that have a low fit score ($<$ 0.5). We can also notice that all attributes that have high accuracy on MCScript also have a high fit score.
\subsection{Implicit Fine-Tune Method}

We develop a method of fine-tuning with additional data based on the deficiencies found in the previous section. We fine-tune on additional data, but we select only data related to attributes that BERT is deficient in. 
 
\subsubsection{Data Selection}
In our experiments, we use RACE \cite{Lai2017RACELR} as our supplementary dataset.
While we can fine-tune on the entire dataset, we can also select a subset that directly targets the deficient attributes in semantic norm.
To select such a subset, we define a datum as related if any words match between the datum in the supplementary dataset and the deficient feature in semantic norm, stemming all words beforehand. For some attributes, we remove frequent words (``is”, ``does", and ``has") to avoid matching too many sentences within RACE.

Since each datum in RACE involves a question, answer, and passage, we allow matches between either of the three texts, and do not differentiate between matches in the question, answer, and passage. We find that this keeps around a third of the data in RACE (around 44K, out of the 97K data present in RACE). It is also key that this data selection process does not require access to the downstream task dataset. Thus, this procedure has the ability to generalize to other tasks beyond MCScript 2.0.

\subsubsection{Fine-Tuning Procedure}

We fine-tune BERT's language objectives on RACE. We do not change the properties of either objective, to keep comparability between our analysis and BERT. This mimics \citet{devlin2018}, and thus, we fine-tune the token masking objective and the next sentence prediction objective. Several works have improved on BERT's language objectives \cite{Yang2019XLNetGA, Liu2019RoBERTaAR}, but we keep the language objectives in BERT intact for comparison.

After fine-tuning on RACE, we fine-tune on MCScript with the classification objective only. We do this since we need to build a classification layer for the specific task, as noted in \citet{devlin2018}. We do not freeze the weights in this process, as to keep comparability with the fine-tuning procedure in \citet{devlin2018}.

\subsection{Explicit Fine-Tune Method}

Motivated by our results in \ref{explicit}, we develop a method of integrating knowledge graph embeddings with the BERT embeddings. First, we query knowledge graphs based on the given text to find relationships between objects in the text. Then, we generate an embedding for each relationship found (similar to Section \ref{explicit}). Finally, we fine-tune these embeddings alongside the BERT embeddings.

\subsubsection{Knowledge Graph Query}

\begin{figure*}
    \centering
  \includegraphics[width=\textwidth]{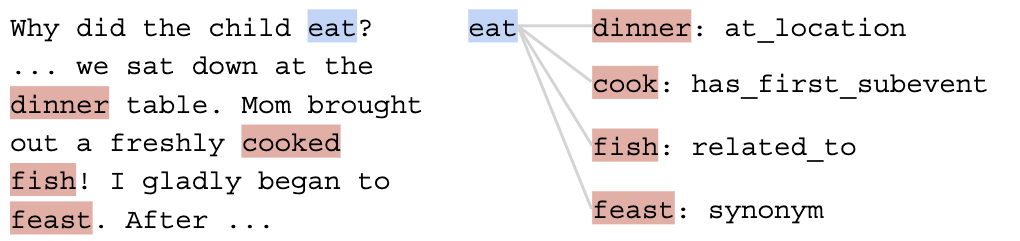}
\caption{Visualization of ConceptNet knowledge base queries. The word $eat$ is being queried with the other words in the text, with the valid edges discovered displayed against the left.}
\label{vis_kb}
\end{figure*}

We query a suite of knowledge bases (ConceptNet \cite{conceptnet}, WebChild \cite{tandon-etal-2017-webchild}, ATOMIC \cite{Sap2018ATOMICA}) to create knowledge graph embeddings. First, we examine all relationships, indexing each unique relationship sequentially. Then, during fine-tuning, for each prompt in MCScript 2.0, we query the knowledge bases to find any \textit{(start\_node, end\_node, edge)} matches between the knowledge base and the current prompt. For example, if $eat$ and $dinner$ are both present in the text, the relationship $at\_location$ in ConceptNet would match (Figure \ref{vis_kb}). We record the index of the matched relationship, keeping a list of matched relationships per word in the prompt. If a \textit{start\_node} spans more than one word, we record the match as occurring for the first word in the phrase.

\subsubsection{Fine-Tuning Procedure}

We fine-tune our knowledge graph embeddings alongside the BERT fine-tuning procedure. We randomly initialize an embedding for each relationship and each knowledge graph. We choose an embedding for each word in the prompt (randomly, if there is more than one relationship associated), creating a sequence of knowledge graph embeddings. We create a sequence embedding for the 30-dimensional graph embeddings by feeding the sequence through an bidirectional LSTM. Then, during fine-tuning, we classify each datum in MCScript based on the concatenation of the explicit graph sequence representation and the BERT sequence embedding (i.e. $[CLS]$), as per \citet{devlin2018}.

\subsection{Results and Analysis}

\begin{table}[]
\centering
\begin{tabular}{ll|l} \hline
System & Acc. & Data \\ \hline
BERT$_{LARGE}$ + RACE & 84.3 & 98 K \\
BERT$_{LARGE}$ + RACE (random) & 84.0 & 44 K \\
BERT$_{LARGE}$ + RACE (selected) & \textbf{84.5} & 44 K \\ \hline
\end{tabular}
\caption{Test set results from the implicit method on MCScript 2.0. ``selected" indicates a subset of RACE that consists of misclassified attributes in semantic norm. ``random" is a randomly chosen subset.}
\label{racetable}
\end{table}

\begin{table}[]
\centering
\begin{tabular}{ll} 
\hline
System & Accuracy \\ \hline
Human \cite{mcscript2} & 97.4 \\
Random Baseline & 48.9 \\ \hline
BERT$_{LARGE}$ & 82.3 \\
with ConceptNet & 83.1 \\
with WebChild & 82.7 \\
with ATOMIC & 82.5 \\
with all KB & 83.3 \\
with all KB + RACE (selected) & \textbf{85.5} \\ \hline
\end{tabular}
\caption{Test set results for knowledge base embeddings on MCScript 2.0.}
\label{explicit_results}
\end{table}

Table \ref{racetable} shows the results from the implicit method. Accuracy is consistent across the board, with all models giving about a 2\% downstream accuracy boost. However, the model with the less amount of data (RACE, selected from deficiencies only) achieves equivalent accuracy to the entire RACE dataset, while using only half the amount of data. This underscores the importance of the abstract semantic norm task, as the related data selection process was effective in choosing examples that are directly related to deficiencies. 

Table \ref{explicit_results} shows our results with explicit knowledge embeddings. Each knowledge base improves accuracy, with ConceptNet giving the largest performance boost. ATOMIC gives the smallest boost, likely because the ATOMIC edges involve longer phrases, which means less matches, and the overlap between ATOMIC text and the text present in the task is not as large as either ConceptNet or WebChild.

We can also combine the explicit knowledge base embeddings and the implicit RACE fine-tuning, yielding the highest accuracy (with all KB + RACE (subset) in Table \ref{explicit_results}). The knowledge embeddings provide a similar +1\% absolute improvement (85.5 vs. 84.5), suggesting that the knowledge embeddings cover different aspects and relationships in the text than learned during fine-tuning on RACE. 

\section{Related Work}

Similar to our attribute classification investigation, several other works have used applied semantic norm datasets to computational linguistics \cite{Agirre2009ASO, Bruni2012DistributionalSI, Kiela2016VirtualEA}. Methodologically, our work is most similar to \citet{lucy2017}, who use a logistic regression classifier to determine fit score of word type embeddings based on leave-one-out verification. \citet{forbes2019neural} investigates the commonsense aptitude of contextual representations. However, our work differs in several important ways: 1) we connect our analysis to downstream reasoning aptitude, underscoring the importance of the semantic norm analysis, and 2) we introduce various ways of improving BERT, motivated by our analysis.

In contemporaneous work, various research has been done in improving upon BERT's performance through knowledge augmentation. Implicitly, \citet{Sun2019HowTF} explores fine-tuning on in-domain data, similarly to our fine-tuning on the RACE dataset \cite{Lai2017RACELR}. They discover an increase in accuracy that is especially prevalent over smaller datasets. Our work differs in that we do not fine-tune on the entire domain data, but rather select a smaller subset of data to fine-tune on. Other work extends BERT to domains where its original training data does not suffice \cite{Beltagy2019SciBERTPC, Lee2019BioBERTAP}. RoBERTa \cite{Liu2019RoBERTaAR} also pretrains on RACE, and finds increased results through altering several of BERT's pretraining tasks, claiming that BERT was extensively undertrained. Explicitly, ERNIE, \citet{Sun2019ERNIEER} introduces information to contextual representations during pretraining. ERNIE uses word-level fusion between the contextual representation and explicit information.

Prior work has developed several benchmark datasets to assess commonsense knowledge of NLP models \cite{Roemmele2011,Mostafazadeh2016ACA,zhang-etal-2017-ordinal,zellers2018swagaf,zellers2019hellaswag,Ostermann2018MCScriptAN,mcscript2,Sakaguchi2019WINOGRANDEAA}.
These benchmarks are typically posed as question answering, but we use semantic norm datasets to specifically assess BERT's ability to represent grounded attributes.
Further, we demonstrate that these abstract attributes can be used to enhance BERT's representations and improve the downstream performance.

\section{Conclusion}

We found that BERT outperforms previous distributional methods on an attribute classification task, highlighting possible reasons why BERT improves the state-of-the-art on various commonsense reasoning tasks. However, we show that BERT still lacks proper attribute representations in many areas.
We developed implicit and explicit methods of remedying this deficit on the downstream task.
We demonstrated that, individually and combined, both methods can improve scores on the downstream reasoning task.
We motivate future work in probing and improving the ability of neural language models to reason about everyday commonsense.

\section*{Acknowledgments}
The authors thank Maxwell Forbes, Keisuke Sakaguchi, and Noah A. Smith as well as the anonymous reviewers for their helpful feedback. JD and JK are  supported by NSF Multimodal and the Funai Overseas Scholarship respectively.

\bibliography{emnlp-ijcnlp-2019}

\begin{thebibliography}{33}
\expandafter\ifx\csname natexlab\endcsname\relax\def\natexlab#1{#1}\fi

\bibitem[{Agirre et~al.(2009)Agirre, Alfonseca, Hall, Kravalova, Pasca, and
  Soroa}]{Agirre2009ASO}
Eneko Agirre, Enrique Alfonseca, Keith~B. Hall, Jana Kravalova, Marius Pasca,
  and Aitor Soroa. 2009.
\newblock A study on similarity and relatedness using distributional and
  wordnet-based approaches.
\newblock In \emph{NAACL-HLT}.

\bibitem[{Beltagy et~al.(2019)Beltagy, Cohan, and Lo}]{Beltagy2019SciBERTPC}
Iz~Beltagy, Arman Cohan, and Kyle Lo. 2019.
\newblock Scibert: Pretrained contextualized embeddings for scientific text.
\newblock \emph{ArXiv}, abs/1903.10676.

\bibitem[{Bruni et~al.(2012)Bruni, Boleda, Baroni, and
  Tran}]{Bruni2012DistributionalSI}
Elia Bruni, Gemma Boleda, Marco Baroni, and Nam-Khanh Tran. 2012.
\newblock Distributional semantics in technicolor.
\newblock In \emph{Proc. of ACL}.

\bibitem[{Davis and Marcus(2015)}]{Davis2015CommonsenseRA}
Ernest Davis and Gary Marcus. 2015.
\newblock Commonsense reasoning and commonsense knowledge in artificial
  intelligence.
\newblock \emph{Commun.\ ACM}, 58.

\bibitem[{Devereux et~al.(2014)Devereux, Tyler, Geertzen, and
  Randall}]{Devereux2014}
Barry~J. Devereux, Lorraine~K. Tyler, Jeroen Geertzen, and Billi Randall. 2014.
\newblock \href {https://doi.org/10.3758/s13428-013-0420-4} {The centre for
  speech, language and the brain ({CSLB}) concept property norms}.
\newblock \emph{Behavior Research Methods}, 46(4).

\bibitem[{Devlin et~al.(2019)Devlin, Chang, Lee, and Toutanova}]{devlin2018}
Jacob Devlin, Ming-Wei Chang, Kenton Lee, and Kristina Toutanova. 2019.
\newblock \href {https://arxiv.org/abs/810.04805} {{BERT}: Pre-training of deep
  bidirectional transformers for language understanding}.
\newblock In \emph{Proc. of NAACL-HLT}.

\bibitem[{Forbes et~al.(2019)Forbes, Holtzman, and Choi}]{forbes2019neural}
Maxwell Forbes, Ari Holtzman, and Yejin Choi. 2019.
\newblock Do neural language representations learn physical commonsense?
\newblock \emph{Proc. of the 41st Annual Conference of the Cognitive Science
  Society}.

\bibitem[{Gordon and van Durme(2013)}]{Gordon2013ReportingBA}
Jonathan Gordon and Benjamin van Durme. 2013.
\newblock Reporting bias and knowledge acquisition.
\newblock In \emph{Proc. of AKBC}.

\bibitem[{Kiela et~al.(2016)Kiela, Bulat, Vero, and Clark}]{Kiela2016VirtualEA}
Douwe Kiela, Luana Bulat, Anita~L. Vero, and Stephen Clark. 2016.
\newblock Virtual embodiment: A scalable long-term strategy for artificial
  intelligence research.
\newblock \emph{ArXiv}, abs/1610.07432.

\bibitem[{Lai et~al.(2017)Lai, Xie, Liu, Yang, and Hovy}]{Lai2017RACELR}
Guokun Lai, Qizhe Xie, Hanxiao Liu, Yiming Yang, and Eduard~H. Hovy. 2017.
\newblock {RACE}: Large-scale reading comprehension dataset from examinations.
\newblock In \emph{Proc. of EMNLP}.

\bibitem[{Lee et~al.(2019)Lee, Yoon, Kim, Kim, Kim, So, and
  Kang}]{Lee2019BioBERTAP}
Jinhyuk Lee, Wonjin Yoon, Sungdong Kim, Donghyeon Kim, Sunkyu Kim, Chan~Ho So,
  and Jaewoo Kang. 2019.
\newblock Biobert: a pre-trained biomedical language representation model for
  biomedical text mining.
\newblock \emph{ArXiv}, abs/1901.08746.

\bibitem[{Liu et~al.(2019{\natexlab{a}})Liu, Gardner, Belinkov, Peters, and
  Smith}]{Liu2019LinguisticKA}
Nelson~F. Liu, Matthew~Ph Gardner, Yonatan Belinkov, Matthew~E. Peters, and
  Noah~A. Smith. 2019{\natexlab{a}}.
\newblock Linguistic knowledge and transferability of contextual
  representations.
\newblock In \emph{Proc. of NAACL-HLT}.

\bibitem[{Liu et~al.(2019{\natexlab{b}})Liu, Ott, Goyal, Du, Joshi, Chen, Levy,
  Lewis, Zettlemoyer, and Stoyanov}]{Liu2019RoBERTaAR}
Yinhan Liu, Myle Ott, Naman Goyal, Jingfei Du, Mandar~S. Joshi, Danqi Chen,
  Omer Levy, Miranda Paige~Linscott Lewis, Luke~S. Zettlemoyer, and Veselin
  Stoyanov. 2019{\natexlab{b}}.
\newblock Roberta: A robustly optimized bert pretraining approach.
\newblock \emph{ArXiv}, abs/1907.11692.

\bibitem[{Lucy and Gauthier(2017)}]{lucy2017}
Li~Lucy and Jon Gauthier. 2017.
\newblock \href {https://arxiv.org/abs/1705.11168} {Are distributional
  representations ready for the real world? {E}valuating word vectors for
  grounded perceptual meaning}.
\newblock In \emph{Proc. of RoboNLP}.

\bibitem[{McCarthy(1960)}]{mcc}
Jeanette McCarthy. 1960.
\newblock Programs with common sense.

\bibitem[{Mostafazadeh et~al.(2016)Mostafazadeh, Chambers, He, Parikh, Batra,
  Vanderwende, Kohli, and Allen}]{Mostafazadeh2016ACA}
Nasrin Mostafazadeh, Nathanael Chambers, Xiaodong He, Devi Parikh, Dhruv Batra,
  Lucy Vanderwende, Pushmeet Kohli, and James~F. Allen. 2016.
\newblock A corpus and cloze evaluation for deeper understanding of commonsense
  stories.
\newblock In \emph{Proc. of NAACL-HLT}.

\bibitem[{Ostermann et~al.(2018)Ostermann, Modi, Roth, Thater, and
  Pinkal}]{Ostermann2018MCScriptAN}
Simon Ostermann, Ashutosh Modi, Michael Roth, Stefan Thater, and Manfred
  Pinkal. 2018.
\newblock \href {https://www.aclweb.org/anthology/L18-1564} {{MCS}cript: A
  novel dataset for assessing machine comprehension using script knowledge}.
\newblock In \emph{Proc. of LREC}, Miyazaki, Japan. European Languages
  Resources Association (ELRA).

\bibitem[{Ostermann et~al.(2019)Ostermann, Roth, and Pinkal}]{mcscript2}
Simon Ostermann, Michael Roth, and Manfred Pinkal. 2019.
\newblock \href {https://doi.org/10.18653/v1/S19-1012} {{MCS}cript2.0: A
  machine comprehension corpus focused on script events and participants}.
\newblock In \emph{Proc. of *{SEM}}, Minneapolis, Minnesota. Association for
  Computational Linguistics.

\bibitem[{Pennington et~al.(2014)Pennington, Socher, and
  Manning}]{pennington-etal-2014-glove}
Jeffrey Pennington, Richard Socher, and Christopher Manning. 2014.
\newblock \href {https://doi.org/10.3115/v1/D14-1162} {{G}lo{V}e: Global
  vectors for word representation}.
\newblock In \emph{Proc. of EMNLP}, Doha, Qatar. Association for Computational
  Linguistics.

\bibitem[{Peters et~al.(2018)Peters, Neumann, Iyyer, Gardner, Clark, Lee, and
  Zettlemoyer}]{Peters2018}
Matthew Peters, Mark Neumann, Mohit Iyyer, Matt Gardner, Christopher Clark,
  Kenton Lee, and Luke Zettlemoyer. 2018.
\newblock \href {http://www.aclweb.org/anthology/N18-1202} {Deep contextualized
  word representations}.
\newblock In \emph{Proc. of NAACL-HLT}.

\bibitem[{Radford et~al.(2018)Radford, Narasimhan, Salimans, and
  Sutskever}]{openaigpt}
Alec Radford, Karthik Narasimhan, Tim Salimans, and Ilya Sutskever. 2018.
\newblock \href
  {https://s3-us-west-2.amazonaws.com/openai-assets/research-covers/language-unsupervised/language_understanding_paper.pdf}
  {Improving language understanding by generative pre-training}.

\bibitem[{Roemmele et~al.(2011)Roemmele, Bejan, and Gordon}]{Roemmele2011}
Melissa Roemmele, Cosmin~Adrian Bejan, and Andrew~S. Gordon. 2011.
\newblock Choice of plausible alternatives: An evaluation of commonsense causal
  reasoning.
\newblock In \emph{AAAI 2011 Spring Symposium}.

\bibitem[{Sakaguchi et~al.(2019)Sakaguchi, Bras, Bhagavatula, and
  Choi}]{Sakaguchi2019WINOGRANDEAA}
Keisuke Sakaguchi, Ronan~Le Bras, Chandra Bhagavatula, and Yejin Choi. 2019.
\newblock {WINOGRANDE}: An adversarial winograd schema challenge at scale.
\newblock \emph{ArXiv}, abs/1907.10641.

\bibitem[{Sap et~al.(2019)Sap, Bras, Allaway, Bhagavatula, Lourie, Rashkin,
  Roof, Smith, and Choi}]{Sap2018ATOMICA}
Maarten Sap, Ronan~Le Bras, Emily Allaway, Chandra Bhagavatula, Nicholas
  Lourie, Hannah Rashkin, Brendan Roof, Noah~A. Smith, and Yejin Choi. 2019.
\newblock \href {https://arxiv.org/abs/1811.00146} {{ATOMIC}: An atlas of
  machine commonsense for if-then reasoning}.
\newblock In \emph{Proc. of AAAI}.

\bibitem[{Schubert(2015)}]{Schubert2015WhatKO}
Lenhart Schubert. 2015.
\newblock What kinds of knowledge are needed for genuine understanding?
\newblock In \emph{Proc of Cognitum}.

\bibitem[{Speer and Havasi(2013)}]{conceptnet}
R.~Speer and Catherine Havasi. 2013.
\newblock Conceptnet 5: A large semantic network for relational knowledge.
\newblock In \emph{The People's Web Meets NLP}.

\bibitem[{Sun et~al.(2019)Sun, Qiu, Xu, and Huang}]{Sun2019HowTF}
Chi Sun, Xipeng Qiu, Yige Xu, and Xuanjing Huang. 2019.
\newblock How to fine-tune {BERT} for text classification?
\newblock \emph{ArXiv}, abs/1905.05583.

\bibitem[{Tandon et~al.(2017)Tandon, de~Melo, and
  Weikum}]{tandon-etal-2017-webchild}
Niket Tandon, Gerard de~Melo, and Gerhard Weikum. 2017.
\newblock \href {https://www.aclweb.org/anthology/P17-4020} {{W}eb{C}hild 2.0 :
  Fine-grained commonsense knowledge distillation}.
\newblock In \emph{Proc. of {ACL} 2017, System Demonstrations}, Vancouver,
  Canada. Association for Computational Linguistics.

\bibitem[{Yang et~al.(2019)Yang, Dai, Yang, Carbonell, Salakhutdinov, and
  Le}]{Yang2019XLNetGA}
Zhilin Yang, Zihang Dai, Yiming Yang, Jaime~G. Carbonell, Ruslan Salakhutdinov,
  and Quoc~V. Le. 2019.
\newblock {XLNet}: Generalized autoregressive pretraining for language
  understanding.
\newblock \emph{ArXiv}, abs/1906.08237.

\bibitem[{Zellers et~al.(2018)Zellers, Bisk, Schwartz, and
  Choi}]{zellers2018swagaf}
Rowan Zellers, Yonatan Bisk, Roy Schwartz, and Yejin Choi. 2018.
\newblock {SWAG}: A large-scale adversarial dataset for grounded commonsense
  inference.
\newblock In \emph{Proc. of EMNLP}.

\bibitem[{Zellers et~al.(2019)Zellers, Holtzman, Bisk, Farhadi, and
  Choi}]{zellers2019hellaswag}
Rowan Zellers, Ari Holtzman, Yonatan Bisk, Ali Farhadi, and Yejin Choi. 2019.
\newblock \href {https://arxiv.org/abs/1905.07830} {{HellaSwag}: Can a machine
  really finish your sentence?}
\newblock In \emph{Proc. of ACL}.

\bibitem[{Zhang et~al.(2017)Zhang, Rudinger, Duh, and
  Van~Durme}]{zhang-etal-2017-ordinal}
Sheng Zhang, Rachel Rudinger, Kevin Duh, and Benjamin Van~Durme. 2017.
\newblock \href {https://doi.org/10.1162/tacl_a_00068} {Ordinal common-sense
  inference}.
\newblock \emph{TACL}, 5.

\bibitem[{Zhang et~al.(2019)Zhang, Han, Liu, Jiang, Sun, and
  Liu}]{Sun2019ERNIEER}
Zhengyan Zhang, Xu~Han, Zhiyuan Liu, Xin Jiang, Maosong Sun, and Qun Liu. 2019.
\newblock \href {https://www.aclweb.org/anthology/P19-1139} {{ERNIE}: Enhanced
  language representation with informative entities}.
\newblock In \emph{Proc. of ACL}, Florence, Italy. Association for
  Computational Linguistics.

\end{thebibliography}
\bibliographystyle{acl_natbib}

\newpage

\appendix
\section{Appendices}
\subsection{Hyperparameters}
Seen in Table \ref{er-hyp} is a list of hyperparameters for our experiments. We use the same parameters for both uses of explicit knowledge embeddings. 

\begin{table}[h]
\small
\centering
\begin{tabular}{ |l c|}
\hline
\multicolumn{2}{|c|}{Regression Classifier}\\
Penalty & L2\\
\# Penalty Coefficient & 1.0\\
Iteration count & 200\\
Optimizer & lbfgs \\
Patience & 1e-4 \\
\multicolumn{2}{|c|}{Explicit Knowledge Embeddings}\\
Embedding size & 10 \\
Knowledge bases used & 3 \\
\multicolumn{2}{|c|}{BERT Fine-Tuning}\\
Maximum sequence length & 450 \\
Train batch size & 32 \\
Learning rate & 1e-5 \\
Epochs & 4 \\
Warmup & 20\% \\
\multicolumn{2}{|c|}{LSTM}\\
Hidden size & 32 \\
Dropout & 0.0 \\
Bidirectional & Yes \\
\hline
\end{tabular}
\caption{Hyperparameters used throughout experiments.}
\label{er-hyp}
\end{table}

\subsection{Contextualization Module Pseudocode}

Psuedocode can be found by referencing Algorithm 1.

\begin{algorithm*}
  \SetAlgoLined
  \DontPrintSemicolon
  \textbf{contextualize} \textit{(object, attribute):}\;
    to\_remove = [does]\;
    \If{attribute[first word] in to\_remove}{
        attribute[second word] = make\_plural(attribute[second word])\;
        attribute.remove(attribute[first word])\;
    }
    \eIf{starts\_with\_vowel(attribute[first word])}{
        $c_{\text{prefix}}$ = An
    }{
        $c_{\text{prefix}}$ = A
    }
    needs\_affix = [made]\;
    \eIf{attribute[first word] in needs\_affix}{
        $c_{\text{affix}}$ = is
    }{
        $c_{\text{affix}}$ = None
    }
    $c_{\text{postfix}}$ = .\;
    return $c_{\text{prefix}}$ + object + $c_{\text{affix}}$ + attribute + $c_{\text{postfix}}$
  \caption{Contextualization Module for CSLB Attributes}
\end{algorithm*}
\end{document}